\documentclass[10pt,twocolumn,letterpaper]{article}

\usepackage[pagenumbers]{cvpr}      

\usepackage{graphicx}
\usepackage{amsmath}
\usepackage{amssymb}
\usepackage{booktabs}
\usepackage{setspace}
\usepackage{multirow} 
\usepackage[pagebackref,breaklinks,colorlinks]{hyperref}

\usepackage{xcolor}
\usepackage{ulem}

\newcommand{\yl}[1]{{\color{black}#1}}
\newcommand{\zwy}[1]{{\color{black}#1}}
\newcommand{\zwyy}[1]{{\color{black}#1}}

\newcommand{\jkw}[1]{{\color{black}{#1}}}

\usepackage[capitalize]{cleveref}
\crefname{section}{Sec.}{Secs.}
\Crefname{section}{Section}{Sections}
\Crefname{table}{Table}{Tables}
\crefname{table}{Tab.}{Tabs.}

\def\cvprPaperID{236} 
\def\confName{CVPR}
\def\confYear{2023}

\newcommand{\papername}{LC-NeRF}

\begin{document}

\title{\papername: Local Controllable  Face Generation in Neural Randiance Field}


\author{
Wenyang Zhou$^{1}$ \qquad Lu Yuan$^{2}$ \qquad Shuyu Chen$^{3}$ \qquad Lin Gao$^{3}$ \qquad Shimin Hu$^{1}$ \vspace{5pt} \\
$^{1}$Tsinghua University \qquad
$^{2}$Stanford University \qquad \\
$^{3}$Institute of Computing Technology, CAS and University of Chinese Academy of Sciences \qquad
}

\twocolumn[{%
\renewcommand\twocolumn[1][]{#1}%
\maketitle
\begin{center}
\centering
\captionsetup{type=figure}
\vspace{-7mm}
  \includegraphics[width=1.\linewidth]{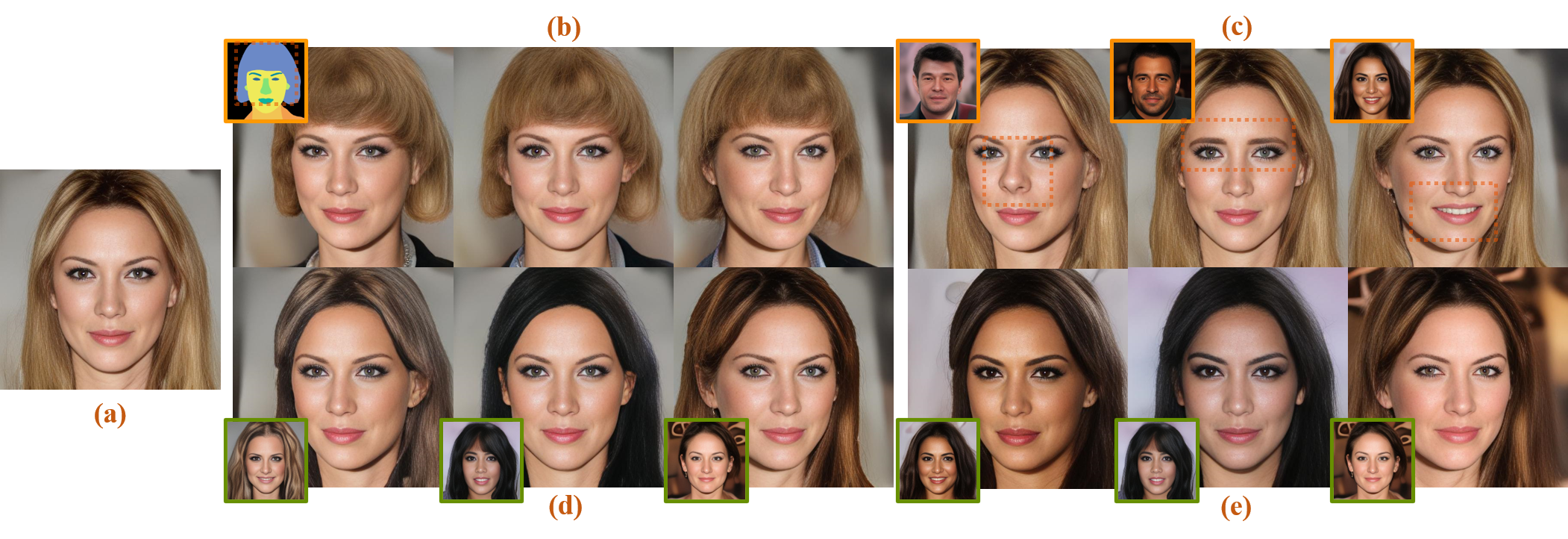}
  \caption{Given an input image (a) and a given reference face image, our method can independently edit the geometry of any local region\zwyy{,} such as hair, nose, mouth, eyebrows, etc. We show these local \zwyy{and global} face editing tasks achieved with our method: (b) \zwyy{editing} the geometry of the hair by modifying the semantic mask while retaining the geometry of other regions and 3D consistency; (c) \zwyy{editing} the geometry of more local regions, such as nose, eyebrows and mouth; (d) \zwyy{editing} the local texture of the hair while retaining the geometry and 3D consistency; (e) \zwyy{editing} the global texture while retaining the geometry and 3D consistency.}
 \label{fig:teaser}
\end{center}%
}]


\begin{abstract}
3D face generation has achieved high visual quality and 3D consistency thanks to the development of neural radiance fields (NeRF). Recently, to generate and edit 3D faces with NeRF representation, some methods are proposed and achieve good results in decoupling geometry and texture. The latent codes of these generative models affect the whole face, and hence modifications to these codes cause the entire face to change. However, users usually edit a local region when editing faces and do not want other regions to be affected. Since changes to the latent code affect global generation results, these methods do not allow for fine-grained control of local facial regions. To improve local controllability in \zwyy{NeRF-based face} editing, we propose \papername, which is composed of a \textit{Local Region Generators Module} and a \textit{Spatial-Aware Fusion Module}, allowing for local geometry and texture control of local facial regions. Qualitative and quantitative evaluations show that our method provides better local editing than state-of-the-art face editing methods. Our method also performs well in downstream tasks, such as text-driven facial image editing.
\end{abstract}

\section{Introduction}
\label{sec:intro}
Realistic face image generation and editing is a useful topic in image synthesis and is widely used in portrait generation and artistic creation. 
Many efforts\cite{karras2019style,karras2020analyzing,karras2021alias} have been paid to improve the quality and increase the resolution of the generated face images.
At the same time, users want to have more interaction with and control over the generated images.
To increase the controllability of the generation process, many methods are proposed to edit the face images by different interfaces, such as sketches\cite{DeepFaceEditing}, texts\cite{Patashnik_2021_ICCV}, semantic masks\cite{shi2022semanticstylegan}, etc. 



Benefiting from the implicit 3D representation of neural radiance fields 
(NeRF)\cite{mildenhall2020nerf} 
, the image synthesis models have shown significant progress in transferring 2D 
image generation task \cite{karras2019style} to 3D
\cite{eg3d, or2022stylesdf, gu2021stylenerf},
addressing 3D consistency 
in perspective transformation.
EG3D\cite{eg3d}, StyleNeRF\cite{gu2021stylenerf}, and StyleSDF\cite{or2022stylesdf} use implicit three-dimensional representations to improve the quality of 3D face generation.
Recently, some \zwyy{NeRF-based face} editing methods\cite{sun2022fenerf,sun2022ide,10.1145/3550469.3555377} have shown excellent results in decoupling the geometry and texture of faces. FENeRF\cite{sun2022fenerf}, IDE-3D\cite{sun2022ide} and NeRFFaceEditing\cite{10.1145/3550469.3555377} decouple geometry and texture by using separate geometry and texture networks. 
These methods use the global latent code to generate global 3D representation, so \zwyy{editing} the latent code will affect the whole face. This will inevitably affect non-editing regions when editing local facial regions, and even lead to inconsistent facial identities. 

To improve the controllability of \zwyy{NeRF-based face} editing, we propose a local controllable face generation and editing method, named \papername, for fine-grained facial local region control and the decoupling of geometry and texture. 
There are two core issues that need to be solved, one is the decomposition of the global 3D representation and representations of the local 3D regions, and another is the fusion of local 3D regions.
It is challenging to decompose a complete 3D representation into multiple local 3D representations and stably complete the training process. To overcome \zwyy{this} issue, we design our generator network with multiple local generators to generate the content for each local region. In addition, for more flexible control over geometry and texture, we further subdivide the local generator into a geometry network and a texture network controlled by geometry code and texture code separately. Through these designs
, 
our method can modify the geometry and texture of local regions without affecting other regions by \zwyy{editing} multiple local latent codes. 
Another core challenge is how to fuse local 3D representations of all local regions to generate the final face image. We propose a Spatial-Aware Fusion Module to complete the fusion of multiple local regions. Specifically, each local geometry generator predicts the semantic confidence of spatial points, and the fusion module fuses the features of different local generators in a soft and smooth way through all confidences.

Qualitative and quantitative experiments show that our method not only better achieves the stability of non-editing regions during editing, but also better ensures the consistency of face identities than state-of-the-art face editing methods. The main contributions of this paper are summarized as followed:
\begin{itemize}
    \item We propose a local controllable NeRF face generation and editing method, named \papername, to control and edit the geometry or texture of local regions in a decoupled manner.
    \item 
    We propose a \textit{Local Region Generators Module} to decompose the global 3D representation and latent codes into multiple local regions, and a \textit{Spatial-Aware Fusion Module} that aggregates these regions into a whole image.
    \item Our method achieves state-of-the-art 3D local geometry and texture editing results for face editing,  as demonstrated by both qualitative and quantitative evaluations.
\end{itemize}

\section{Related Work}
\label{sec:rw}
\subsection{Neural Face Generation}
Generative models, such as Stylegan v1-v3\cite{karras2019style,karras2020analyzing,karras2021alias}, have achieved high-qulity generaton of 2D images. 
In recent years, NeRF\cite{gropp2020implicit} has emerged as a method that can implicitly model 3D geometry from 2D images and then render photorealistic and 3D consistent images. 
Subsequently, NeRF-based face generative models have been investigated. PI-GAN \cite{chanmonteiro2020pi-GAN} proposes a SIREN-based\cite{sitzmann2020implicit} implicit radiance field to generate 3D faces via sampled latent and positonal encoding. 
\zwy{Furthermore, due to the advantages of StyleGAN\cite{karras2019style} in image generation, some methods\cite{or2022stylesdf,gu2021stylenerf} 
based on StyleGAN can generate high resolution and quality images.}
StyleNeRF~\cite{gu2021stylenerf} provides a 3D GAN approach that fuses style-based generation with scene representation by neural radiance fields. StyleSDF~\cite{or2022stylesdf} is similar, but incorporates an SDF-based 3D representation to ensure that images generated from different viewpoints have 3D geometric consistency. 
In addition, some methods study different forms of space representation. For example, EG3D\cite{eg3d} uses three projection planes (tri-plane) to represent the 3D space, generated by a backbone of StyleGAN. 
GRAM\cite{deng2022gram} proposes a radiance manifolds based generative model that divides the space into multi-manifolds. 
These methods improves the quality of generated images but lack the editability and controllability of geometry and texture. 
Our method enhances the effect of disentanglement of facial features while maintaining generative quality.

\begin{figure*}[t]
  \centering
   \includegraphics[width=0.95\linewidth]{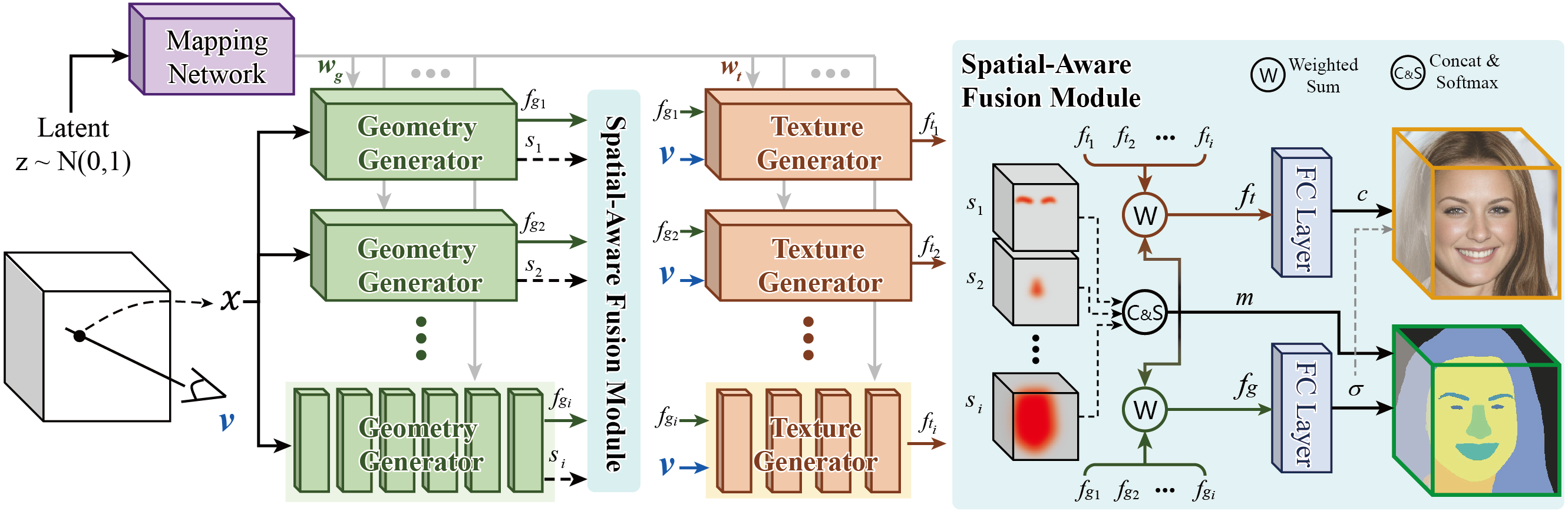}
   \vspace{-2mm}
   \caption{Pipeline of our framework \papername. Our pipeline is composed of multiple local generators and a spatial aware fusion module. The local generators include geometry and texture generators, separately controlled by geometry latent code $w_g$ and texture latent code $w_t$. 
   \papername \ can modify the geometry or texture of an local region directly by \zwyy{editing} its latent code $w_g$ or $w_t$.}
   \label{fig:pipeline}
\end{figure*}

\subsection{Neural Face Editing}
With the high-quality generation of images, many portraits are generated by the generation models, as described above. Meanwhile, some methods\cite{Faceshop} take the editing as an application.
Editing tasks are no longer limited to the 2D domain, and research on how to perform editing and control on 3D faces becomes pupolar. In the image domain, Faceshop\cite{Faceshop} treats the face editing task as sketch-based image completion that can only edit the facial geometry. The demands for face editing are no longer
\zwyy{editing} geometry but also modifying texture, such as \zwyy{editing} hair colors. DeepFaceEditing\cite{DeepFaceEditing} 
decouples facial local regions by using sketches to represent geometry.
SofGAN\cite{chen2022sofgan} trains a semantic occupancy field (SOF) and uses 2D semantic masks to generate face images to decouple geometry and texture. 
SemansticStyleGAN\cite{shi2022semanticstylegan} enhances the control over local regions by generating the features of each region separately and then fusing the features of different regions in the 2D feature domain.
The implicit 3D representation and generation of high-quality multi-view images in NeRF inspire works on 3D face decoupling and editing.

FENeRF\cite{sun2022fenerf} adds a mask branch to PI-GAN \cite{chanmonteiro2020pi-GAN} for geometry control. Further, IDE-3D\cite{sun2022ide} and NeRFFaceEditing\cite{10.1145/3550469.3555377} realize the decoupled control of geometry and texture based on three projection planes\cite{eg3d}. IDE-3D\cite{sun2022ide} proposes a framework with separate geometry and texture networks to generate respective tri-plane features. 
Inspired by AdaIN, NeRFFaceEditing \cite{10.1145/3550469.3555377} decomposes the tri-plane features into geometry features and appearance features for decouping the geometry and appearance.
These methods are all implemented by optimizing the latent code during geometry editing, which is used to generate the whole face. 
Therefore, the global effect is prone to be affected during local editing.

\section{Methodology}
\label{sec:algoritam}

In this section, we introduce the architecture of our method in detail. We aim to control and edit local regions by \zwyy{editing} the local geometry and texture latent codes.
To achieve this goal, we need to solve two core problems: i) How to control the geometry and texture of each local region separately; ii) How to fuse the features of all \yl{the} local regions into the global feature and generate a whole face image. For the first problem, we propose two independent lightweight local networks for each region: a geometry and a texture network, controlled by their respective geometry and texture latent codes (Section \ref{sec:LocalRegionGenerators}). For the second problem, we design a spatial aware fusion module to fuse the features generated by all \yl{the} local networks and then generate the final face image (Section \ref{sec:SpatialFusion}). We introduce two discriminators and detailed loss functions used in network training (Section \ref{sec:Discriminators}). \zwy{Then, we will introduce how to encode the real image to the latent code through GAN inversion and how to perform mask editing (Section \ref{sec:InversionEditing}).}

\subsection{Local Region Generators}
\label{sec:LocalRegionGenerators}

\paragraph{Geometry Generator}
The geometry generator is designed to determine the shape of the face. We assign a lightweight geometry generator $\Phi s_i$ for each local region $i$ of the face. If a 3D point belongs to a certain local region, the corresponding geometry generator provides the most information for this point. The generator plays a major role in determining the semantic category and geometry information of the point. As shown in Figure \ref{fig:pipeline}, each geometry generator contains 6 linear layers with SIREN\cite{sitzmann2020implicit} activation, and is controlled by the geometry latent code $w_g$.

Given a sampled 
point $x \in \mathbb{R}^{3}$, the $i_{th}$ geometry generator module $\Phi s_i$ decodes it to obtain the semantic confidence $s_i(x)$ and geometry feature $f_{g_i}(x)$ from a geometry latent $w_{g_i}$:

\begin{equation}
  s_i(x), f_{g_i}(x) = \Phi s_i(x,w_{g_i})
\label{eq:eq1}
\end{equation}

Here, $s_i(x)$ indicates the probability that the $i_{th}$ local geometry generator believes three-dimensional point $x$ to be in its region. $s_i (x)$ has two characteristics: i) The larger the value of $s_i(x)$, the more importance and more \jkw{proportion} 
the features of this generator acquire in the subsequent fusion module; ii) Sampling or modifying the geometry latent $w_{g_i}$ can increase or reduce the $s_i(x)$ value of the local region $i$, which enables local editing of geometry. Specifically, we use a linear layer
following the geometry feature $f_{g_i}(x)$ to calculate the geometry confidence $s_i(x)$.

\paragraph{Texture Generator}
The texture generator can be interpreted as a shader, which is used to fill the color of the geometry generated by the geometry generator. In other words, the texture generators do not participate in or affect the generation of geometry, and the geometry generator is only used to determine the shape of the face, so as to achieve local region decoupling and geometry/texture decoupling. Each texture generator contains 4 linear layers with SIREN\cite{sitzmann2020implicit} activation, and is controlled by the texture latent code $w_t$.

Given the viewing direction $v \in \mathbb{R}^{3}$, the $i_{th}$ local texture generator module $\Phi_{t_i}$ decodes the texture feature $f_{t_i}(x)$ from a geometry latent $w_{t_i}$:

\begin{equation}
  f_{t_i}(x) = \Phi_{t_i} (f_{g_i}(x), v, w_{t_i})
\label{eq:eq8}
\end{equation}

The texture features predicted by all the local texture generators will be fused in subsequent fusion module, and the final color value of the sampled 3D point $x$ will be predicted.

\begin{figure*}
  \centering
   \includegraphics[width=0.92\linewidth]{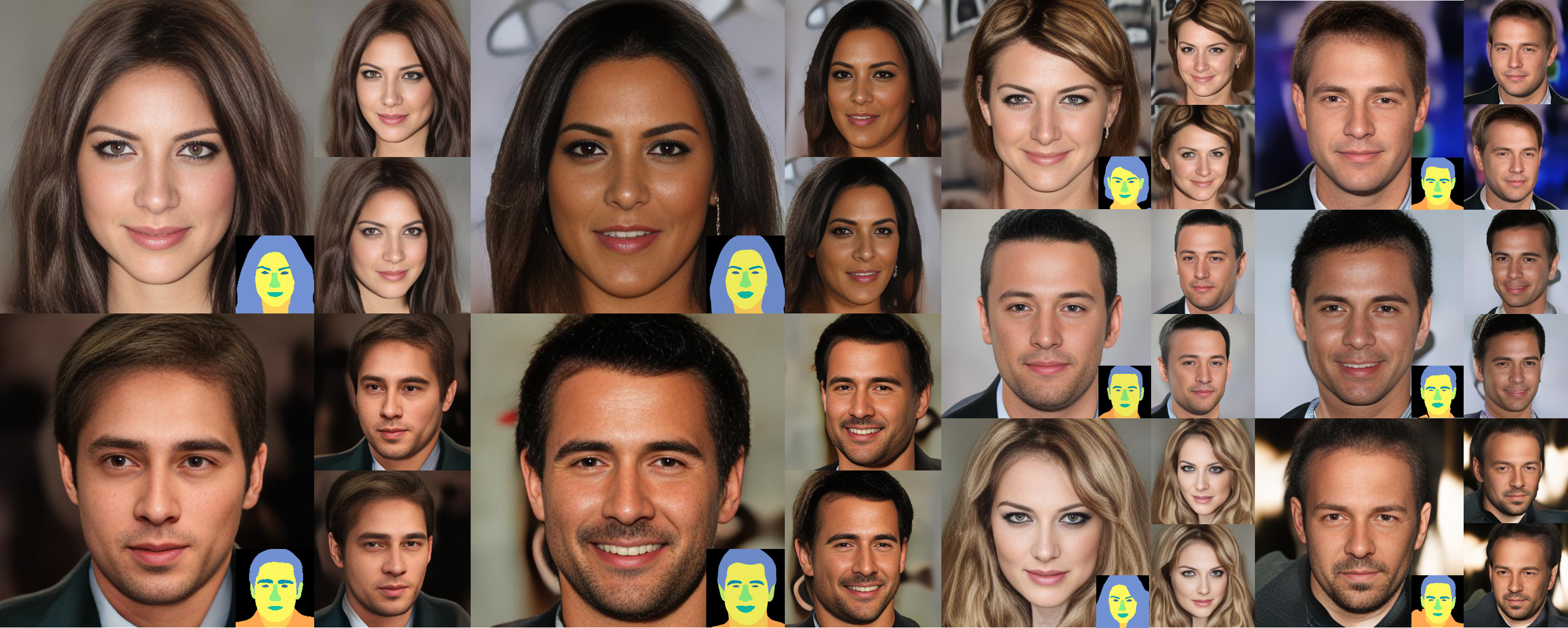}
   \caption{Multi-view 
   face images and semantic masks with a resolution of 512, generated by \papername \ trained on CelebAMask-HQ dataset.}
   \label{fig:generated}
\end{figure*}

\subsection{Spatial-Aware Fusion Module}
\label{sec:SpatialFusion}

The spatial-aware fusion module is designed for interaction and aggregation among multiple local generators. The proposed fusion module fuses the features of different generators with a soft and adjustable mechanism and generates the whole image.
We concatenate the semantic confidence $s_i(x)$ of all \yl{the} geometry generators and apply the softmax activation to obtain the semantic mask $m(x)$.

\begin{equation}
  m_i(x) = \dfrac{{\rm e}^{s_i(x)}} {\sum_{k=1}^K{\rm e}^{s_{c_k}(x)}}
\label{eq:eq4}
\end{equation}

where $K$ is the number of local regions. We use the semantic mask $m(x)$ to fuse the geometry features $f_{g_i}(x)$ to get the final geometry feature $f_g(x)$.  

\begin{equation}
  f_g(x) = \sum_i(m_i(x) * f_{g_i}(x))
\label{eq:eq5}
\end{equation}

$f_g(x)$ is the geometry  feature 
of the 3D point extracted by our proposed geometry generators. We use a linear layer after $f_g(x)$ to predict the signed distance field (SDF) value $d(x)$ of the 3D point $x$. Then we convert the SDF value to volume density $\sigma(x)$ by the following formula\cite{or2022stylesdf}.

\begin{equation}
  \sigma(x) = Sigmoid(d(x)/\beta)/\beta
\label{eq:eq7}
\end{equation}

where $\beta$ is a learnable parameter. The smaller  $\beta$ is, the more the volume density $\sigma(x)$ will converge on the surface of the face. In our experiments, the initial value of $\sigma(x)$ is set to 0.1. As the training progresses, the value of $\beta$ will become smaller and smaller.

The texture features $f_{t_i}(x)$ are also fused with the semantic mask $m(x)$ to get the final texture feature $f_t(x)$. And then we use one linear layer after $f_t(x)$ to get the color value $c(x)$:

\begin{equation}
  f_t(x) = \sum_i(m_i(x) * f_{t_i}(x))
\label{eq:eq9}
\end{equation}

We render the generated image $I^{\prime}$ and the generated semantic mask $M^{\prime}$ through the volume rendering. Given a camera position o, by shooting a ray $r(t)=o+tv$ at each pixel, we calculate the color and mask of N points sampled from $t_n$ to $t_f$ on the ray. In our experiments, N is set to 18.

\begin{equation}
\begin{split}
&I^{\prime}(r)=\int_{t_n}^{t_f} T(t) \sigma(r(t)) c(r(t), v) d t,\\
&M^{\prime}(r)=\int_{t_n}^{t_f} T(t) \sigma(r(t)) m(r(t), v) d t,\\
&\text{where } T(t)=\exp \left(-\int_{t_n}^t \sigma(r(s)) d s\right)
\label{eq:eq7}
\end{split}
\end{equation}

At this point, we complete the fusion operation through spatial aware fusion module to generate the whole image $I^{\prime}$ and the semantic mask $M^{\prime}$ \zwy{with a super resolution model\cite{zhou2022codeformer}}.

\subsection{Discriminators and Loss Function}
\label{sec:Discriminators}

In order to ensure quality of the generated image and correspondence between the image and the mask, we propose a double discriminator supervision strategy. One discriminator is the image quality and pose awared discriminator $D_I$, which is used to distinguish between real images and generated images and predicts the azimuth and the elevation $\theta^{\prime}$. \zwy{In addition to the GAN loss\cite{gan2020}}, we use a smoothed L1 loss $L_{pose}$ and R1 regularization loss to supervise the training of $D_I$ 
for the generated images.

\begin{equation}
\begin{split}
  L_{D_I} &= \mathbb{E}[1+exp(D_I(I^{\prime}))]  + \mathbb{E}[1+exp(-D_I(I))] \\
  &+\lambda_{I_{reg}}\mathbb{E}\|\nabla D_I(I)\|^2  + \lambda_{pose}L_{pose}(\theta, \theta^{\prime})
\label{eq:eq9}
\end{split}
\end{equation}

where $I^\prime$ and $M^\prime$ are the fake image and the semantic mask generated by \papername \ with the sampled pose $\theta$. $I$ and $M$ are the ground truth image and the mask sampled from the real dataset. where $\lambda_{I_{reg}}$, $\lambda_{pose}$ are set to 10 and 15 respectively.

The other discriminator is the image and semantic mask discriminator $D_{IM}$, which is used to determine whether the image is consistent with the semantic mask. We also regularize the gradient norm for this discriminator with R1 regularization loss.

\begin{equation}
\begin{split}
  L_{D_{IM}} &= \mathbb{E}[1+exp(D_{IM}(I^{\prime}, M^{\prime}))]  \\
  &+ \mathbb{E}[1+exp(-D_{I M}(I, M)] \\
  &+ \lambda_{I M_{reg}} \mathbb{E}\| \nabla D_{I M}(I, M )\|^2
\label{eq:eq9}
\end{split}
\end{equation}
where $\lambda_{I M_{reg}}$ is set to 10.

The generator $G$ is supervised by the two discriminators $D_{M}$ and $D_{IM}$ and the camera pose loss $L_{pose}$. In addition, we also introduce geometry supervision of SDF with eikonal loss\cite{gropp2020implicit} and minimal surface loss\cite{or2022stylesdf}.

\begin{equation}
\begin{split}
  L_G &= \mathbb{E}[1+exp(-D_I(I^{\prime}))]  \\
  &+ \lambda_{I M}\mathbb{E}[1+exp(-D_{I M}(I^{\prime}, M^{\prime}))]  \\
  &+ \lambda_{\text { pose }} L_{\text {pose }} (\theta, \theta^{\prime})  + \lambda_{eik} \mathbb{E}[\| \nabla d(x)\|_2 - 1]^2 \\ 
  &+ \lambda_{sur} \mathbb{E}[exp(-100 | d(x)|]
\label{eq:eq10}
\end{split}
\end{equation}

where $\lambda_{I M}$, $\lambda_{pose}$, $\lambda_{eik}$, $\lambda_{sur}$ are set to 0.5, 15, 0.1, 0.05 respectively.


\zwy{
\subsection{Inversion and Editing}
\label{sec:InversionEditing}
We can edit the images generated by the latent codes $w_g$ and $w_s$ at a certain pose as well as the real images. To edit the real images, we need to encode the real images into the $\mathcal{W}^+${\cite{karras2019style}} space through pivotal tuning inversion\cite{roich2021pivotal}. Given a real face image $I$ and the corresponding semantic mask $M$, we first invert $I$ to generate the latent code $w$. When the user edits the mask and gets the edited mask $M_e$, our optimization goal is to find an editing vector $\delta w$ to make the mask $M^\prime$ generated by $\delta w + w$ close to the editing mask $M_e$. 
We use the mean square error (MSE) between the edited mask $M_e$ and generated mask  $M^\prime$. 
During editing, we optimize the geometry latent code of the corresponding local region for 500 iterations.

}

\section{Experiments}
\label{sec:experiments}
In this section, we first introduce our experimental setup, then discuss the generation and comparison results. We present the results of multi-view generation and style transfer of local or global regions. We also discuss the comparison results with state-of-the-art face editing methods, including FENeRF\cite{sun2022fenerf}, IDE-3D\cite{sun2022ide} and NeRFFaceEditing \zwy{(NeRFFE)}\cite{10.1145/3550469.3555377}, to show the superior effectiveness of our \papername.
\paragraph{Training datasets}
We train \papername \ on the CelebAMask-HQ dataset\cite{CelebAMask-HQ}, which contains 30,000 high-quality face images with 1024$\times$1024 resolution. For this dataset, each image provides 
an accurate segmentation image with 19 categories. In our experiments, we combine the left and right local regions into one, such as glasses and eyebrows. After processing, there are 13 types of face local regions.


\begin{figure}
  \centering
   \includegraphics[width=1.\linewidth]{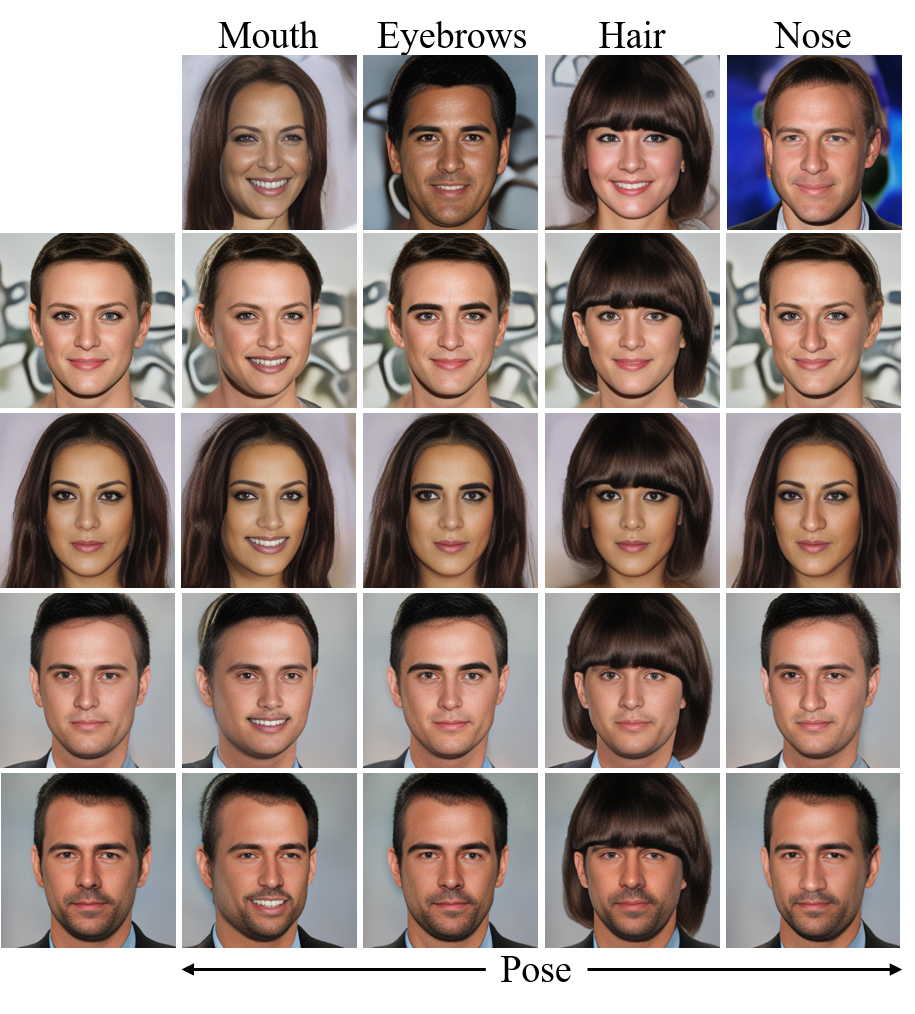}
   \vspace{-7mm}
   \caption{Results of local style transfer. \papername \ supports transferring the geometry and texture of any local region of other faces to the target face. Here we show the results of multi view synthesis that migrate the geometry and texture of an local region at the same time.}
   \label{fig:localstylemix}
\end{figure}

\begin{figure}
  \centering
   \includegraphics[width=1.\linewidth]{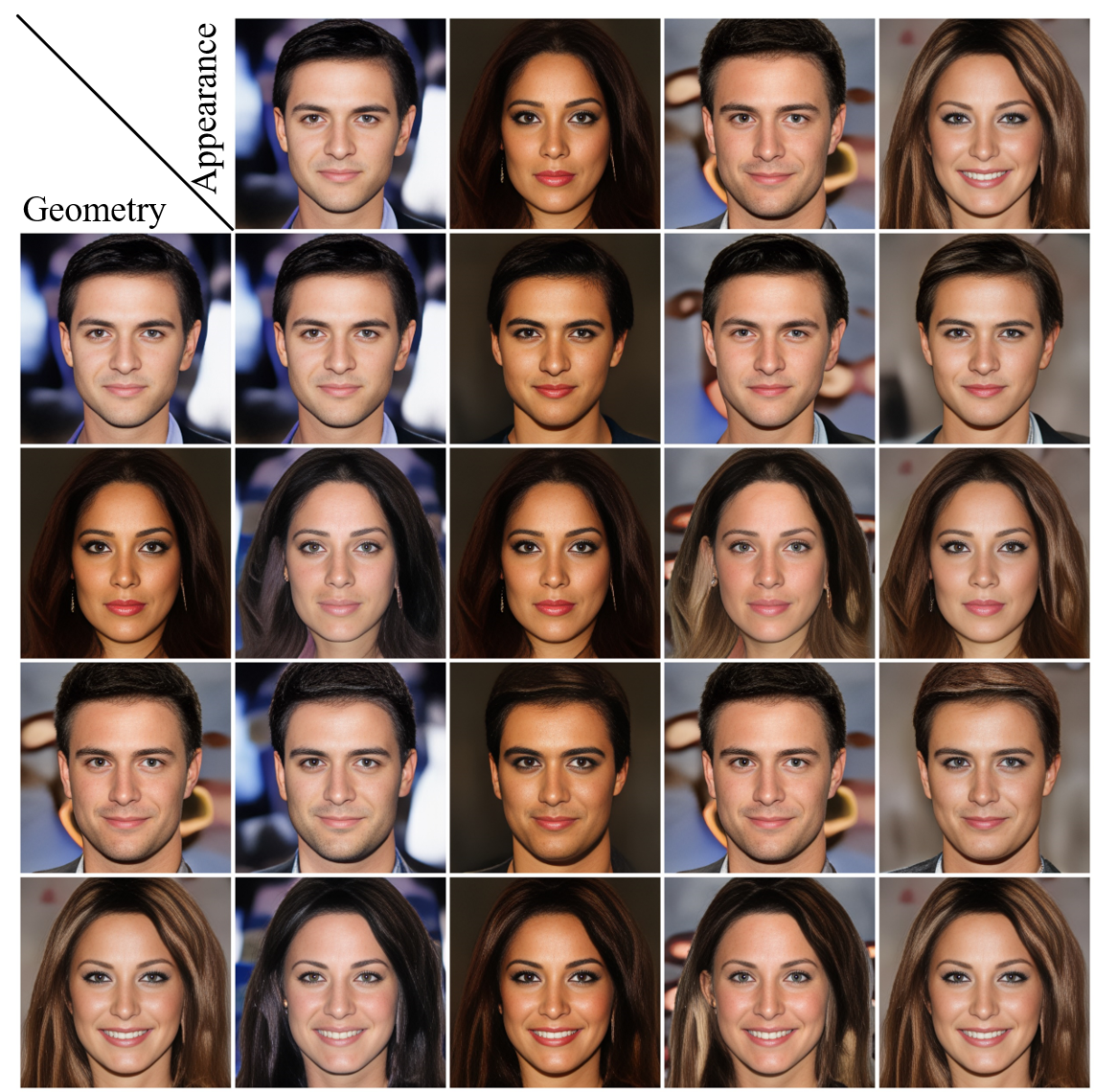}
   \caption{Results of global style transfer. \papername \ supports global modification of face texture. The figure shows examples of transferring the texture information of the reference face to the target face.}
   \label{fig:globalstylemix}
\end{figure}

\begin{table}
  \centering
  \begin{tabular}{@{}lcccc@{}}
    \toprule
    ~ & FENeRF & IDE-3D & NeRFFE & Ours \\
    \midrule 
    Hair  & 0.0332 & 0.0410 & 0.0277 & \textbf{0.0239} \\
    Eyebrow  & 0.0368 & 0.0668 & 0.0188 & \textbf{0.0068} \\
    Nose  & 0.0495 & 0.0538 & 0.0163 & \textbf{0.0078} \\
    Mouth & 0.0539 & 0.0666 & 0.0208 & \textbf{0.0112} \\
    \midrule
    Average  & 0.0433 & 0.0570& 0.0209 & \textbf{0.0124} \\
    \bottomrule
  \end{tabular}
  \caption{Quantitative metrics of pixel level difference in non-editing region for local geometry editing on different local regions. The difference visualization results on the whole image are shown in Figure \ref{fig:GeometryEditing}. }
  \label{tab:example}
\end{table}

\paragraph{Implementation Details}
We use the Adam\cite{kingma2014adam} optimizer with $\beta_1=0$ and $\beta_2=0.9$ to train the generator and discriminators, and the learning rates of $G$, $D_{I}$, $D_{I M}$ are 0.00002, 0.0002 and 0.0002 respectively. \zwy{We train \papername \ on 8 NVIDIA GeForce 3090 GPUs for 48 hours with a batch size of 24. During inference, it takes 0.1s to generate a face image and corresponding semantic mask on 1 NVIDIA GeForce 3090 GPU. \papername \ is implemented on Jittor\cite{hu2020jittor}, which is a fast-training deep learning framework, especially in generating networks\cite{zhou2021jittor}.}

\subsection{Generation Results}
In our framework, we can sample random latent codes 
to generate both face images and semantic masks. As shown in Figure \ref{fig:generated}, our method can generate diverse face images, and the multi-view results prove that our method maintains the 3D consistency across different views. 
In addition, we show the local and global style transfer effects of \papername.

\begin{figure*}[t]
  \centering
   \includegraphics[width=0.9\linewidth]{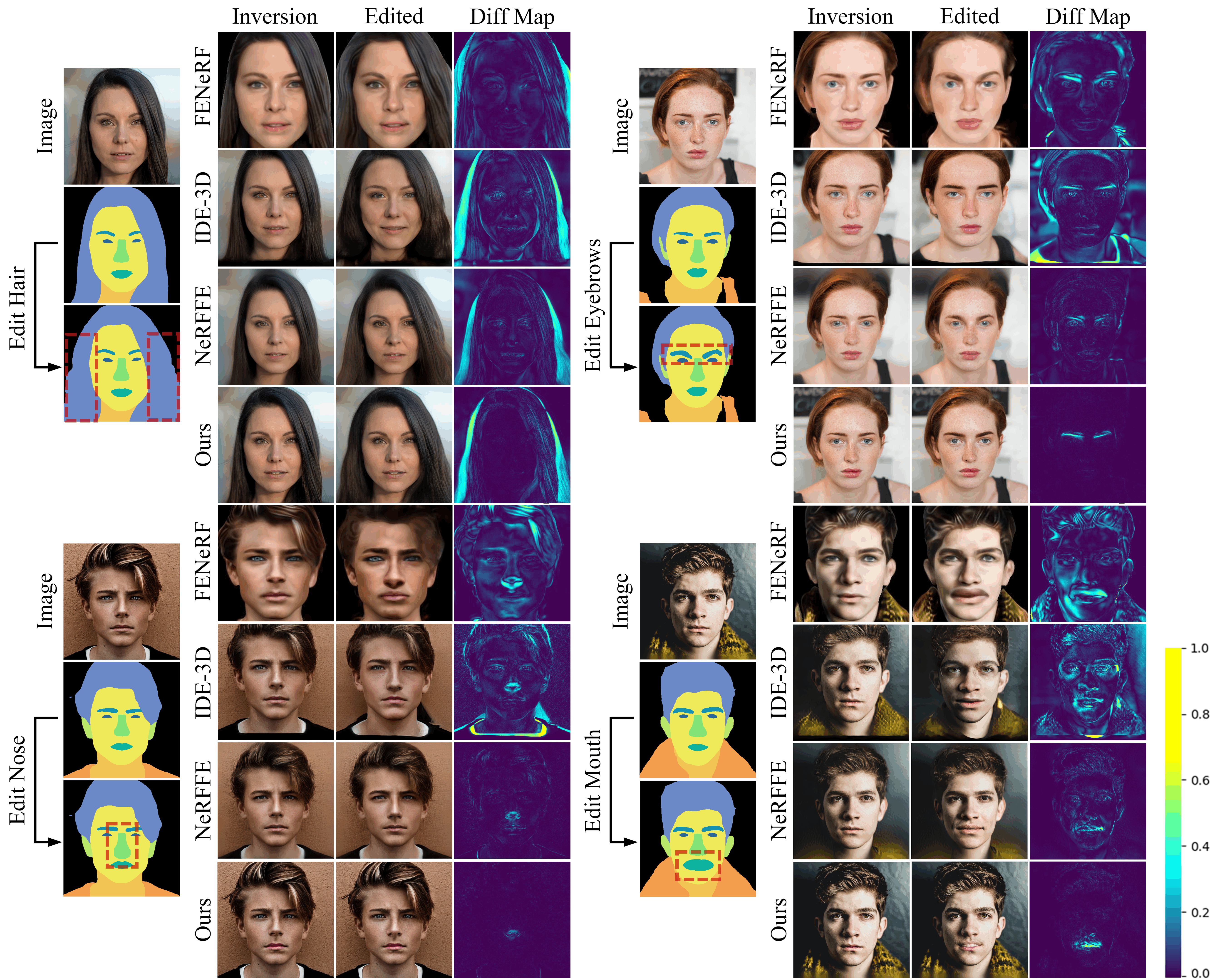}

   \caption{Comparison of local geometry editing with FENeRF\cite{sun2022fenerf}, IDE-3D\cite{sun2022ide}, NeRFFE\cite{10.1145/3550469.3555377}. For each sample, the left side displays the real image, source and edited mask in order from top to bottom. The right side is inversion results, edited results and difference maps of different methods. }
   \label{fig:GeometryEditing}
\end{figure*}

\paragraph{Local style Transfer} \zwy{We can transfer the geometry and texture of local regions.}
For any given face image, we can modify the geometry of specific regions. We directly modify the geometry latent code $w_g$ of an local region to complete local geometry editing. Figure \ref{fig:localstylemix} shows the multi view editing results of modifying mouth, eyebrows, hair, nose and. It can be observed that \papername \ can edit target regions accurately without affecting non-editing regions.

\paragraph{Global style Transfer} We can transfer the geometry and texture of the face globally. We can modify the global texture of all \yl{the} local regions while keeping the geometry unchanged. The global texture can be edited by directly modifying the texture latent codes $w_t$ of all \yl{the} local regions. \zwy{The same goes for modifying global geometry.}
In Figure \ref{fig:globalstylemix}, we show an example of transferring styles of reference images to target images. It can be observed that the geometry of all \yl{the} local regions can remain unchanged when the texture is modified, which also verifies the decoupling property of geometry and texture.

\subsection{Evaluation}

The most important quality of face editing is to change the target region while ensuring that the non-editing regions are not affected. Otherwise, the edited image may become too dissimilar to the original that it may be interpreted as another person entirely. For fair comparison, all \yl{evaluated} methods are tested on the II2S\cite{zhu2020improved} dataset without any fine-tuning. The II2S dataset contains 120 high-quality face images with different styles. We use a pretrained face parsing method\cite{yu2021bisenet} to extract the same semantic mask for all methods.

\begin{figure}[ht]

    \centering
     \includegraphics[width=1
     \linewidth]{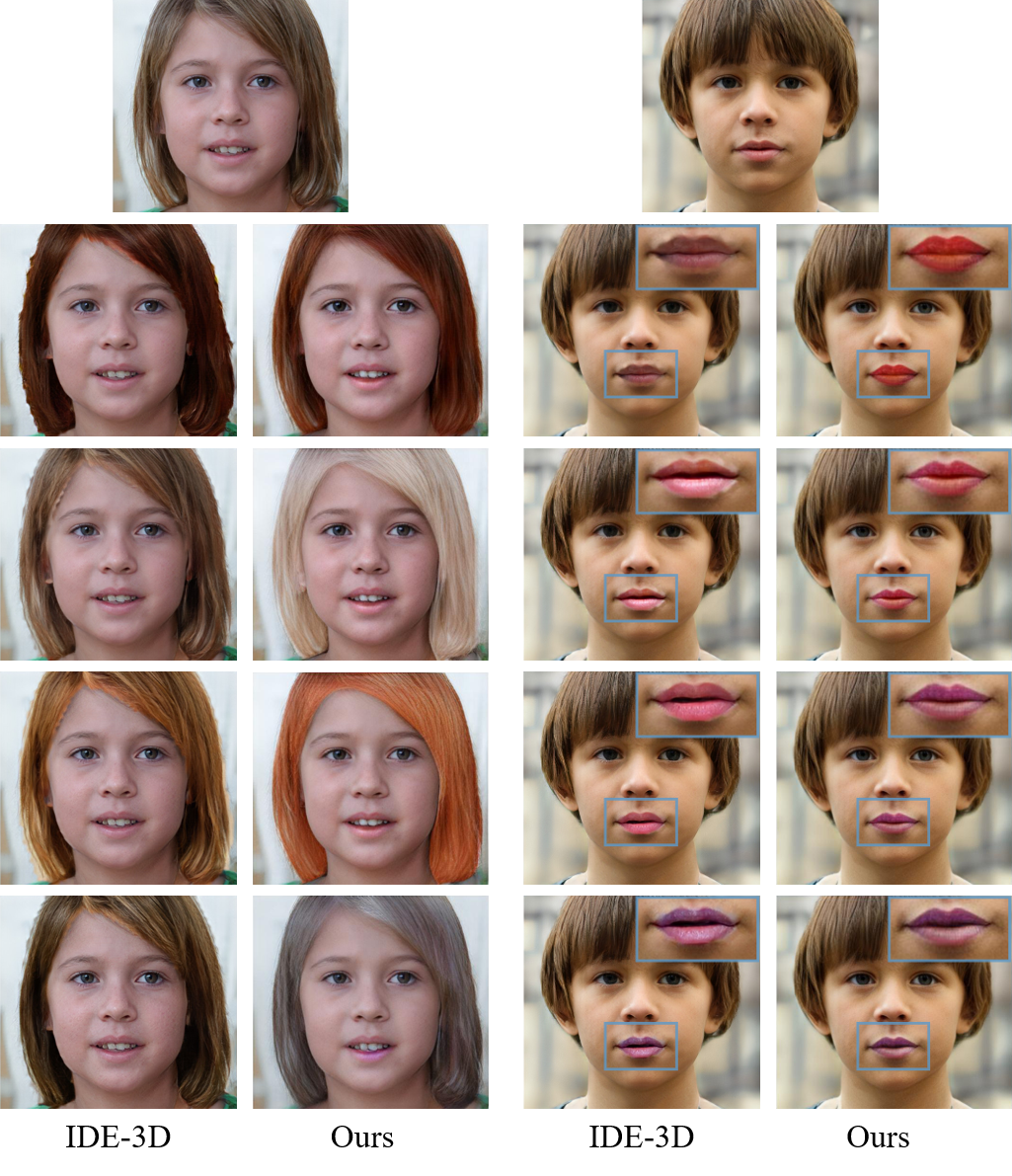}
   \vspace*{-7mm}
   \caption{Qualitative comparison results of real image local texture editing between IDE-3D and \papername(Ours). Two cases show the editing effects of two methods to modify different hair and lip texture colors.
   }
   \label{fig:TextureEditing}
\end{figure}

\paragraph{Real Image Local Geometry Editing} 
Local geometry editing is an interactive and practical application of editing face images by modifying corresponding masks.
In comparison, we appropriately increase the learning rate of IDE-3D inversion to ensure that it converges to the best effect. 
The inversion and local editing results are shown in Figure \ref{fig:GeometryEditing}. Thanks to its local decoupling characteristics, our method \papername \ makes sure that the target regions are modified appropriately, while non-editing regions are not affected. On the other hand, the editing results of FENeRF appear unnatural and unrealistic. Moreover, since IDE-3D and NeRFFaceEditing edit images in the global latent space, the edits inevitably also affect non-editing regions, resulting in obvious changes outside of the target region. In the case of editing the mouth, there are obvious modifications made to other regions of the face in the IDE-3D results. FENERF is limited in hair editing because the face occupies most of the image area during inversion. \zwy{Because NeRFFaceEditing uses VGG loss to explicitly constrain the image invariance of non-editing regions in local editing optimization, there is a relatively good consistency of non-editing regions. However, our method can achieve the best results without such explicit constraints when 
editing local regions. }

We use quantitative metric, i.e., pixel error of non-editing region to evaluate the effectiveness of editing. The pixel error maps $abs(I - I_e)$ of the source image $I$ and the edited image $I_e$ for each editing operation are also visualized in Figure \ref{fig:GeometryEditing}.  The average values of error maps of non-editing region edited locally by different local regions are shown in Table \ref{tab:example}. It can be seen that \papername \ has the highest editing fidelity with lowest image difference value. We also conduct a usability study to evaluate image quality, editing accuracy, and the consistency of non-editing regions. Please refer to the supplementary for more details.

\paragraph{Real Image Local Texture Editing}
Local texture editing allows users to modify the texture of an local region, which emphasizes naturality and harmony. 
\zwy{FENeRF and NeRFFaceEditing are designed for local geometry editing and global texture editing, and do not support local texture editing. So here we compare \yl{the} local texture editing \yl{results} with IDE-3D.}
IDE-3D achieves local texture editing through extracting features from the two triplane features and combining them according to a mask to generate a new face image. This approach makes the generated images unnatural and there is a sense of splicing between different regions. \papername \ can directly change the texture latent code $w_t$ of the certain local region and then fuse the edited high-dimensional texture features, so that the generated image is more natural and controllable. Because authors of IDE-3D has not released their code for local texture editing, we invert the texture editing images given in their papers, which are not real human face images, but images generated by IDE-3D. Local texture editing results are shown in Figure \ref{fig:TextureEditing}. It can be seen that IDE-3D hair texture editing results have a strong sense of border and contain jagged parts. In addition, after editing the mouth texture with IDE-3D, the geometry of the mouth is changed. Some of the results contain closed mouths, while others show  open mouths. This shows that IDE-3D does not achieve effective decoupling of geometry and texture. At the same time, the edited mouth texture is unnatural and foggy and even spreads to non-mouth regions. The images edited by \papername \ are more natural and controllable, benefiting from our proposed local generators and high-dimensional feature fusion mechanism.

\begin{figure}
  \centering
   \includegraphics[width=1.\linewidth]{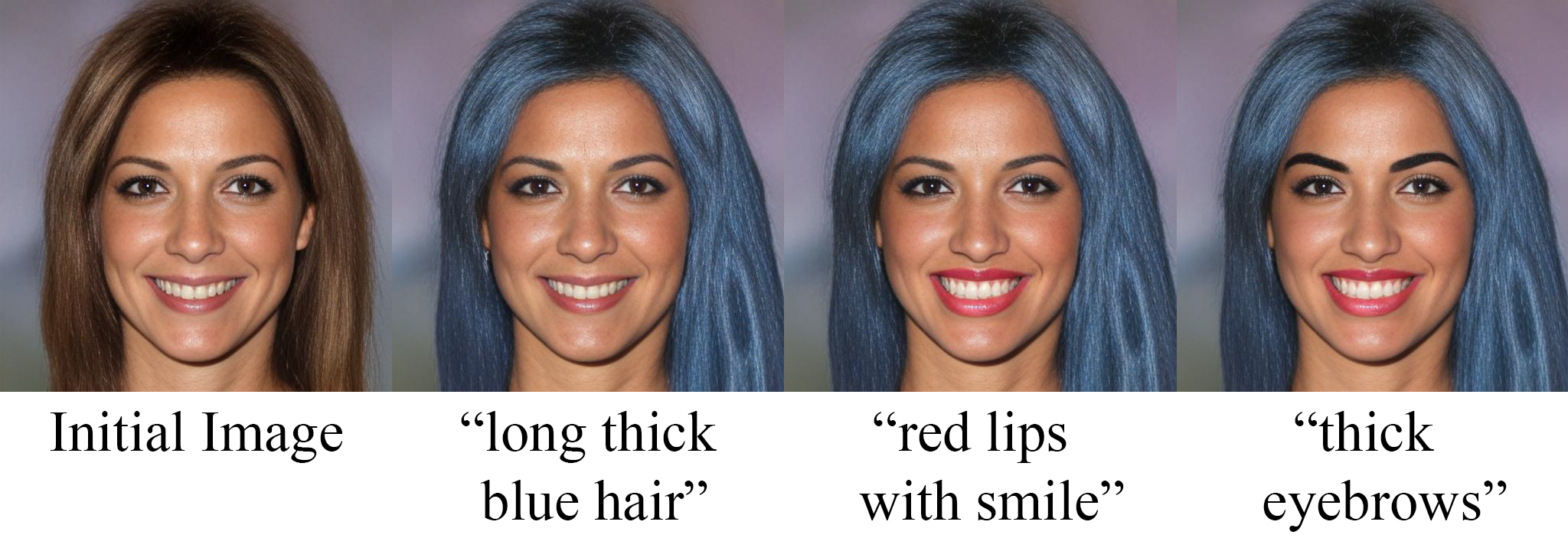}
   \includegraphics[width=\linewidth]{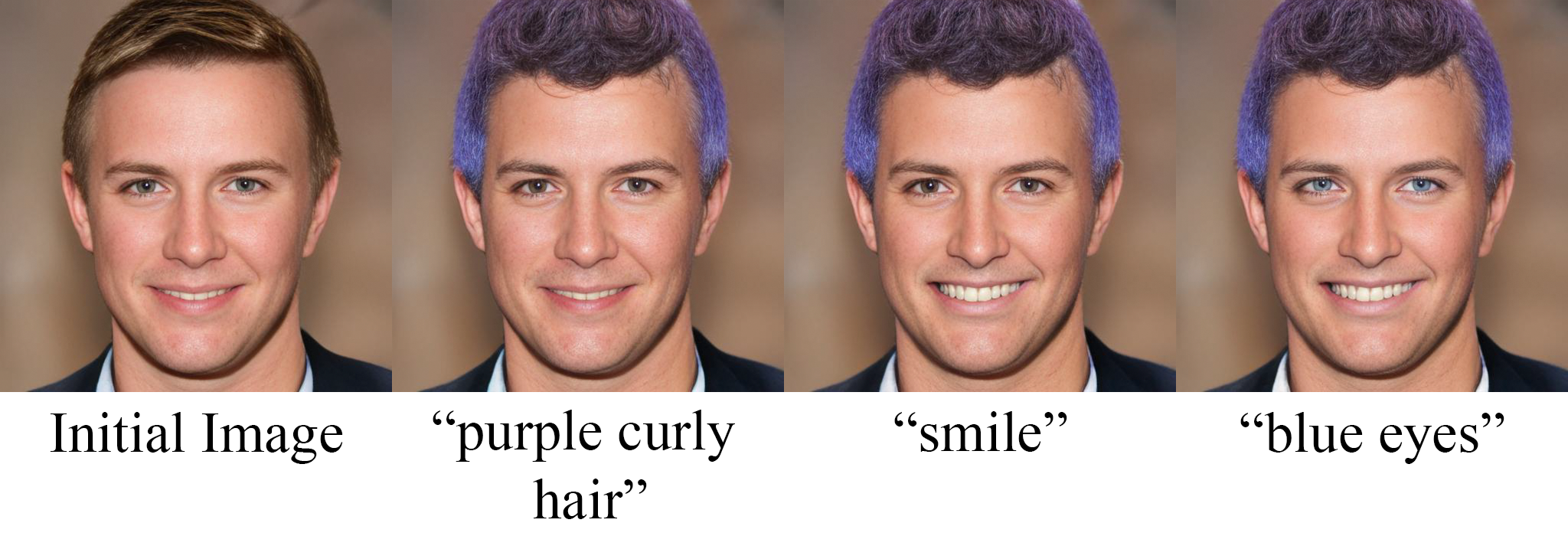}
   \vspace{-7mm}
   \caption{Results of text-driven face editing. Give an initial image (left), \papername \ can edit it directly through the text. The figure shows the results of multiple local region edits, accumulative from left to right.}
   \label{fig:text}
\end{figure}

\subsection{Text-Driven Face Editing}

Text-driven face editing allows users to edit face directly using text, which is an effective and convenient way of editing. Therefore, we also explore 
applicaiton of \papername \ in text-driven image editing. We used StyleCLIP\cite{Patashnik_2021_ICCV} with ViT-B/32 pretrained model for text guided latent manipulation. 
The driving text can directly optimize the $\mathcal{W}^+$ space latent code. In our experiments, generated images are controlled by short text clips, such as "thick eyebrows" and "red lips with smile", using CLIP loss\cite{Patashnik_2021_ICCV}. We present sample edited images with corresponding prompt texts, with 100-300 latent optimizaiton steps, in Figure \ref{fig:text}. In each line of Figure \ref{fig:text}, editing results are accumulated, with each image using the optmized latent from the previous image as a starting point.  
The result shows that \papername \ allows for fine-grained control of facial features and accurate editing driven by text, enabling text-based editing of one facial feature without affecting other regions.

\section{Conclusions\zwy{, Limitations\yl{, a}nd Future \yl{W}ork}}
\label{sec:conclusions}

We propose \papername, a local controllable and editable face generation method, which can generate view-consistent face images and semantic masks.  
Compared with the previous state-of-the-art face editing methods, \papername \ has achieved more fine-grained feature decoupling, including local region decoupling and decoupling of geometry and texture . Our method achieves the best performance in face editing, which ensures the stability of non-editing regions and the consistency of face identities. 
Our method supports local mask editing, local and global texture editing, and can easily be extended to downstream tasks, such as text editing.

\zwy{The limitation of this work is that we can decouple the local regions and the geometry and texture, but we cannot control the local internal texture more finely, such as the hair texture, facial wrinkles, etc. In the future, how to control the content of local texture more finely will be one of our research directions.}

{\small
\bibliographystyle{ieee_fullname}
\bibliography{egbib}
}

\clearpage
\appendix

\section{Comparison of Real Image Geometry Editing}
\label{sec:moredemo}

In this section, we show results of real image geometry editing of single or multiple regions with our method LC-NeRF and three state-of-the-art methods, namely FENeRF\cite{sun2022fenerf}, IDE-3D\cite{sun2022ide}, and NeRFFaceEditing (NeRFFE)\cite{10.1145/3550469.3555377}. The qualitative results are shown in Figure \ref{fig:all1} and Figure \ref{fig:all2}. It can be seen that our method modifies the images more accurately. 

We also use two quantitative methods to evaluate the editing results. One is the average Pixel Difference (PD) between non-editing regions of the original and the edited images to measure whether the editing affects these regions. The other is Mask Consistency (MC) between the editing masks and the masks extracted from the edited images with a pretrained face parser\cite{yu2021bisenet} to measure whether the methods achieve correct and adequate editing. The results are shown in the Table \ref{tab:example}. LC-NeRF achieves the best performance.




\begin{table}[h]
  \centering
  \begin{tabular}{@{}lcccc@{}}
    \toprule
    ~ & FENeRF & IDE-3D & NeRFFE & Ours \\
    \midrule 
    PD  & 0.0441 & 0.0550 & 0.0235 & \textbf{0.0211} \\
    MC  &  0.1826 & 0.1073 & 0.0529 & \textbf{0.0306} \\
    \bottomrule
  \end{tabular}
  \caption{Quantitative comparison of face mask editing of four methods.}
  \label{tab:example}
\end{table}


\begin{figure*}
  \centering
   \includegraphics[width=1\linewidth]{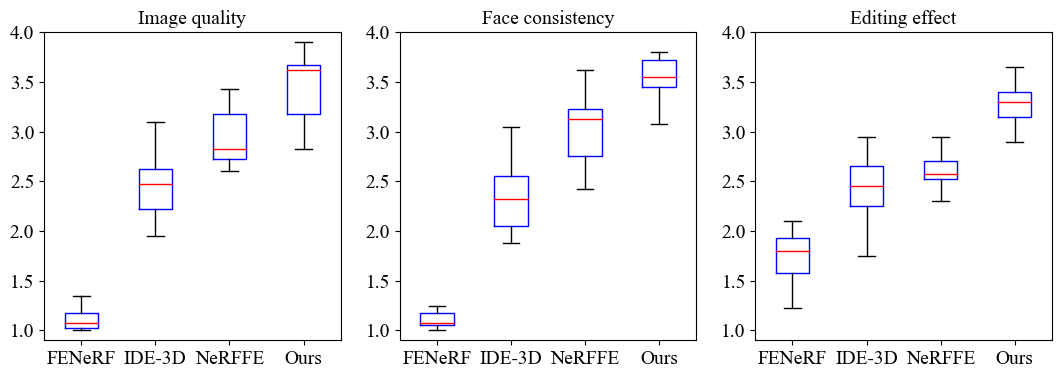}
   \caption{Box plots of image quality, face consistency, and editing effect, based on the participants in the perception study with four methods: FENeRF\cite{sun2022fenerf}, IDE-3D\cite{sun2022ide}, NeRFFE\cite{10.1145/3550469.3555377} and Ours.}
   \label{fig:userstudy}
\end{figure*}

\section{Perception Study}
\label{sec:userstudy}

\begin{figure*}
  \centering
   \includegraphics[width=1\linewidth]{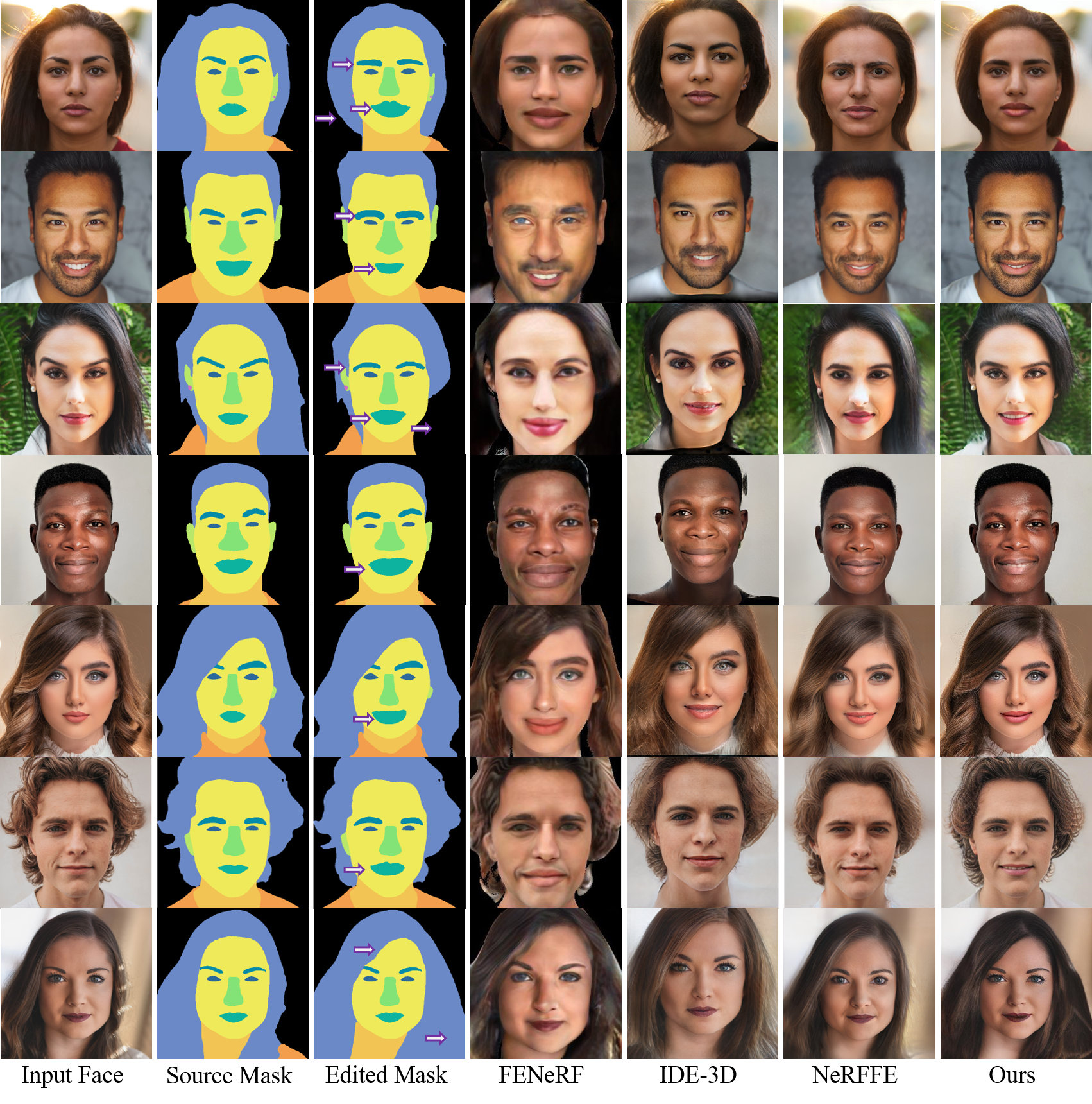}
   \caption{In each case, we show the input face, the source mask, the edited mask (the arrow marks the editing region), and the face editing results of the four methods.}
   \label{fig:all1}
\end{figure*}

\begin{figure*}
  \centering
   \includegraphics[width=1\linewidth]{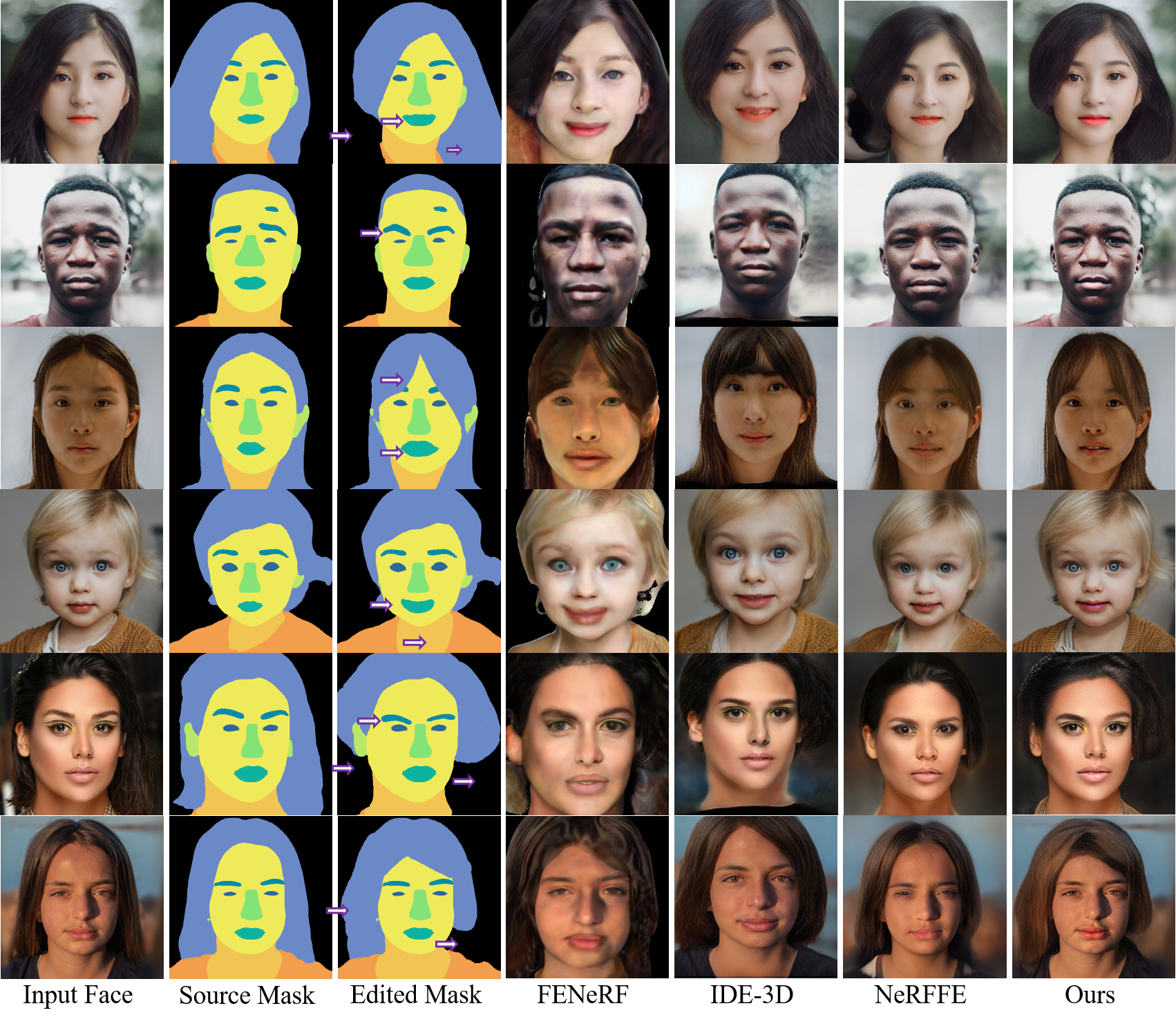}
   \caption{In each case, we show the input face, the source mask, the edited mask (the arrow marks the editing region), and the face editing results of the four methods.}
   \label{fig:all2}
\end{figure*}

For the application of face editing, we conduct a perception study to evaluate the effectiveness. 
We show the input image, the corresponding source mask and the edited mask, and four edited results (including FENeRF\cite{sun2022fenerf}, IDE-3D\cite{sun2022ide}, NeRFFaceEditing (NeRFFE)\cite{10.1145/3550469.3555377}, and Ours) for each editing sample. 
Results from all four methods are placed in random order. 
In this perception study, each participant needs to rank images on 17 examples (Figure \ref{fig:all1} and Figure \ref{fig:all2} in this material as well as Figure 6 in the draft) in the following three aspects respectively:
\begin{itemize}
\setlength{\itemsep}{3pt}
\setlength{\parsep}{0pt}
\setlength{\parskip}{0pt}
    \item \textbf{Image quality}, which measures the quality of face images generated by different methods;
    \item \textbf{Face consistency}, which measures whether the edited face is consistent with the input face, that is, whether the face identity changes before and after editing;
    \item \textbf{Editing effect}, which measures the correctness and the quality of editing results.

\end{itemize}

In total, 40 participants participated in our perception study, and we got 40(participants)$\times$ 17(questions) $=$ 680 subjective evluation for each method.
On average, each researcher spent 21.85 minutes on our survey. When computing the final score, the method ranked as first in each evaluation result translates to a score of four, the second translates to a score of three, the third a score of two, and the last a score of one. 
The results are shown in the Figure \ref{fig:userstudy} in the form of a boxplot.
Our method achieves scores of 3.49, 3.51,  and 3.27 in image quality, face consistency, and editing results respectively, exceeding the scores of FENeRF, IDE-3D, and NeRFFE.
We also perform the ANOVA tests on the three aspects and get the F vlaues for image quality ($F=252.39,p<0.001$), face consistency ($F=215.42,p<0.001$), and editing effect ($F=90.95,p<0.001$). It can be clearly seen that our method has a significant improvement over the other three methods.


\end{document}